\documentclass[10pt,twocolumn,letterpaper]{article}

\usepackage{iccv}
\usepackage{times}
\usepackage{epsfig}
\usepackage{graphicx}
\usepackage{amsmath}
\usepackage{amssymb}
\usepackage{nicefrac}
\usepackage{booktabs}
\usepackage{multirow,multicol}
\usepackage[dvipsnames]{xcolor}
\usepackage{subfig}
\usepackage{pifont}

\usepackage[pagebackref=true,breaklinks=true,letterpaper=true,colorlinks,bookmarks=false]{hyperref}

\iccvfinalcopy %

\def\iccvPaperID{66} %

\ificcvfinal\fi

\begin{document}

\title{Window-based Model Averaging Improves Generalization \\ in Heterogeneous Federated Learning}

\author{Debora Caldarola$^1$\\\
\and
Barbara Caputo$^{1}$\\
$^1$Politecnico di Torino\\
{\tt\small name.surname@polito.it}
\and
Marco Ciccone$^1$
}

\maketitle
\ificcvfinal\thispagestyle{empty}\fi

\begin{abstract}
   Federated Learning (FL) aims to learn a global model from distributed users while protecting their privacy. %
   However, when data are distributed heterogeneously %
   the learning process becomes noisy, unstable, and biased towards the last seen clients' data, slowing down convergence. To address these issues and improve the robustness and generalization capabilities of the global model, we propose \ours\ (\textbf{\longours}). \ours\ aggregates global models from different rounds using a window-based approach, effectively capturing knowledge from multiple users and reducing the bias from the last ones. By adopting a windowed view on the rounds, \ours\ can be applied from the initial stages of training. Importantly, our method introduces no additional communication or client-side computation overhead. Our experiments demonstrate the robustness of \ours\ against distribution shifts and bad client sampling, resulting in smoother and more stable learning trends.
   Additionally, \ours\ can be easily integrated with state-of-the-art algorithms. We extensively evaluate our approach on standard FL benchmarks, %
   demonstrating its effectiveness.    
\end{abstract}

\section{Introduction}

Federated Learning (FL) \cite{mcmahan2017communication} is a distributed machine learning framework aiming at learning a shared global model from edge users' data (the \emph{clients}) while ensuring their privacy. Instead of centrally collecting their data, federated training is based on the exchange of model parameters between clients and the server. The actual training is performed on the client side, and the updates are later aggregated on the server side. In real-world scenarios, the number of clients typically reaches billions \cite{kairouz2021advances}, and their data collection depends on numerous factors such as geographical location \cite{hsu2020federated,fantauzzo2022feddrive,shenaj2023learning,miao2023fedseg}, or personal habits \cite{fallah2020personalized,yang2018applied}. For instance, autonomous vehicles may collect images and videos of largely different cities with varying weather and light conditions \cite{fantauzzo2022feddrive,shenaj2023learning}. This results in highly diverse local data distributions, creating inherent \textit{statistical heterogeneity} within the context of FL \cite{li2020federated,kairouz2021advances}. As a consequence, training a global model capable of addressing the overall underlying distribution becomes particularly challenging: as only a fraction of clients participates in each round, the convergence speed is drastically reduced \cite{li2019convergence,karimireddy2020mime}, the learning trend becomes noisy and unstable \cite{karimireddy2020scaffold,caldarola2022improving},  the clients' biased updates drift the model from its convergence points \cite{karimireddy2020scaffold,li2020federated_fedprox,acar2021federated}, and the global model suffers from \textit{catastrophic forgetting}~\cite{kirkpatrick2017overcoming,caldarola2022improving}, resulting in the loss of knowledge acquired from previous users as training progresses. Most of the approaches addressing these issues focus on client-side training: several methods~\cite{li2020federated_fedprox,karimireddy2020scaffold,acar2021federated,ozfatura2021fedadc,varno2022adabest} regularize the local objective to reduce the \textit{client drift}, while others leverage momentum to incorporate knowledge from previous updates and lead the local optimization onto the path defined by the global models across rounds~\cite{karimireddy2020mime,wang2019slowmo,xu2021fedcm,kim2022communication,liu2023enhance}. More theoretical studies reveal that learning rate decay is fundamental in local training to reach global convergence in heterogeneous settings \cite{li2019convergence,yu2019parallel,cho2022towards,li2023revisiting}. %
Building upon \cite{foret2020sharpness,kwon2021asam}, another promising research direction focuses on the sharpness of reached minima as an indicator of the model's generalization ability, and explicitly guides the local updates towards flatter minima \cite{caldarola2022improving,qu2022generalized,sun2023fedspeed,sun2023dynamic}. Less attention has been given to server-side aggregation. The de-facto standard approach for merging models is FedAvg~\cite{mcmahan2017communication}, where the updated parameters are averaged based on the number of samples seen by each client. Recent studies \cite{hsu2019measuring,reddi2020adaptive} reveal that FedAvg aligns with a step in the optimization path defined by SGD (Stochastic Gradient Descent) \cite{ruder2016overview} with a unitary learning rate, and suggest that using server-side momentum or adaptive optimizers could be beneficial when dealing with heterogeneous scenarios. Differently, \cite{caldarola2022improving} introduces Stochastic Weight Averaging (SWA) \cite{izmailov2018averaging} to ensemble global models across rounds in the later stages of training with the goal of improving stability and generalization. A significant limitation of this approach lies in its impracticality during the early stages of training, rendering it challenging to deploy in real-world contexts. 

In addition, less attention has been given to research in FL related to vision domains \cite{caldarola2022improving}, as most research focuses on its optimization and security aspects. 

In this paper, we aim at building a robust and stable global model without incurring in additional communication or client-side computational burden, with a specific focus on vision tasks. Building upon the insights of \cite{caldarola2022improving}, we propose \textbf{\longours} (\ours), a method for aggregating global models from the initial stages of training. In particular, \ours\ leverages a server-side window-based approach that averages the last $W$ global models. This strategy helps to mitigate the drift introduced by the last seen clients, preserving information from previous users with reduced forgetting. The model built with \ours\ is robust towards both distribution shifts and bad client sampling and can be easily applied on top of any existing state-of-the-art FL algorithm.

Our main contributions are summarized as follows:
\begin{itemize}
    \item We propose \ours\, which averages the last $W$ global models on the server side, building a model more robust towards distribution shifts and bad client sampling from the earliest stages of training.
    \item We show that averaging these models is equivalent to using learning rate decay in the server-side aggregation process.
    \item We evaluate \ours's performances on multiple FL datasets and observe smoother learning trends. Furthermore, we show that the use of \ours\ helps narrow the gap with runs that involve higher client participation rates.
\end{itemize}
\section{Related works}
\paragraph{Federated settings} FL \cite{mcmahan2017communication} enables the training of a shared model among edge devices or institutions while ensuring the privacy of their sensitive data. Real-world scenarios comprise cross-silo and cross-device FL \cite{kairouz2021advances}. The former involves silos like companies or hospitals in the training process, with access to extensive data from multiple clients (\eg, patients). In contrast, the cross-device scenario utilizes billions of edge devices, such as smartphones, which possess limited data and computational resources. Moreover, their data is often \textit{biased} towards various distributions, influenced by factors such as capturing devices, personal habits, and geographical locations \cite{hsu2020federated,kairouz2021advances,fantauzzo2022feddrive,shenaj2023learning,fallah2020personalized}. Lastly, the devices are not always online and reachable, resulting in only a fraction of them available for training. Thus, it is essential to account for constraints related to resource limitations, communication capabilities, and small skewed datasets when designing federated algorithms \cite{li2020federated,kairouz2021advances}. In this work, we focus on the cross-device setting, aiming to avoid adding complexity for the resource-constrained clients while improving the robustness of the global model.

\paragraph{Heterogeneity in Federated Learning} Federated training is based on communication rounds, during which clients and server exchange the global model updated parameters, with the server never accessing the local data. On the server side, the updates are aggregated, usually using a weighted average as introduced by the de-facto standard FedAvg \cite{mcmahan2017communication}. While being effective in homogeneous scenarios, FedAvg fails at achieving comparable performances in heterogeneous cross-device ones \cite{kairouz2021advances,li2020federated}. In particular, local different distributions lead to the so-called \textit{client drifts} \cite{karimireddy2020scaffold}, \ie the local models converging towards different solutions in the loss landscape, making server-side aggregation more challenging. As a consequence, convergence is slowed down \cite{karimireddy2020scaffold,li2019convergence},  the learning trend becomes noisy and unstable \cite{caldarola2022improving} and the global model suffers from catastrophic forgetting of the knowledge acquired by previously involved clients \cite{kirkpatrick2017overcoming,shoham2019overcoming}. As a first step to overcome these issues, \cite{reddi2020adaptive} explains that applying FedAvg on the global updates is equivalent to globally using SGD with learning rate 1, and shows that adaptive optimizers can help address heterogeneous scenarios more effectively. 
To reduce the client drift, FedProx \cite{li2020federated_fedprox} introduces a regularization term in the local objectives, while SCAFFOLD \cite{karimireddy2020scaffold} leverages stochastic variance reduction \cite{reddi2016stochastic} and FedDyn \cite{acar2021federated} aligns local and global stationary points at convergence. However, \cite{varno2022adabest} shows that FedDyn is often prone to parameter explosion in particularly skewed and cross-device settings, introducing AdaBest as a solution. Other approaches use a momentum term \cite{sutskever2013importance} to preserve the history of the previous updates and reduce the bias towards the last fraction of selected clients \cite{wang2019slowmo,karimireddy2020mime,liu2023enhance}. In particular, FedAvgM \cite{hsu2019measuring} uses momentum on the server-side aggregation, while FedACG \cite{kim2022communication} and FedCM \cite{xu2021fedcm} leverage a momentum term to guide local updates in the direction followed by global models. MIME \cite{karimireddy2020mime} combines both stochastic variance reduction and momentum so that local updates mimic the behavior of training on \emph{i.i.d.} data. Since our approach only looks at the server-side aggregation of global models, it can be easily combined with any of these methods. Furthermore, \cite{li2019convergence,yu2019parallel,cho2022towards,li2023revisiting} highlight that employing learning rate decay in the client-side training is essential to achieve convergence in heterogeneous scenarios. In this study, we illustrate that averaging the model parameters at different rounds is equivalent to applying global decay in the SGD steps of FedAvg, resulting in notable improvements in stability during the training process. Lastly, the authors in \cite{luo2021no} reveal that the classifier is the network component most affected by local distribution shifts. %
In this context, our work demonstrates how \ours\ improves the backbone's ability to extract better features, consequently enhancing the stability of the classifier.

Building upon \cite{foret2020sharpness,kwon2021asam,izmailov2018averaging}, other works look at the generalization of the global model through the lens of the loss landscape, linking it with convergence to flat minima. FedSAM \cite{caldarola2022improving,qu2022generalized} uses Sharpness-Aware Minimization (SAM) \cite{foret2020sharpness} optimizing both loss value and sharpness, and FedSpeed \cite{sun2023fedspeed} follows a similar approach, while FedSMOO \cite{sun2023dynamic} introduces the concept of global sharpness. Following this line of research, \cite{caldarola2022improving} shows that using Stochastic Weight Averaging (SWA) \cite{izmailov2018averaging} on the server-side to ensemble global models leads to more robust and stable results with significant gains in performances and generalization. SWA achieves this by averaging the weights obtained by SGD during its optimization path, utilizing a cyclic learning schedule to explore broader regions in the weight space. However, it's important to note that SWA can only be effectively applied near convergence; otherwise, it may hinder the training process. In this work, we address this last issue emerging with SWA by leveraging a window-based approach. Instead of collecting \textit{all} global models from the beginning, \ours\ only averages the last $k$ ones, resulting in significant improvements and overcoming the issues faced with SWA.

\paragraph{Model ensembling for robustness} Our work is inspired by research conducted outside the federated scenario, which demonstrates the effectiveness of model ensembling in enhancing accuracy and robustness. We leverage these valuable insights to face challenges proper of the heterogeneous federated scenarios, aiming at improving the performance of the learned models. In their work, ~\cite{lakshminarayanan2017simple} demonstrate that ensembling predictions in the output space can lead to performance boosts due to the diversity of the networks. Perhaps surprisingly, \cite{garipov2018loss} reveals that randomly initialized networks independently trained on the same task are connected by simple curves of low-loss, and proposes FGE (Fast Geometric Ensembling) to ensemble predictions at the end of weight space exploration. %
Finally, \cite{izmailov2018averaging} shows that solutions found by FGE are found on the edge of the most desirable ones, and presents SWA to ensemble models in the weight space and move towards the center of the minimum. However, SWA is most effective near convergence, \eg after 75\% of training was performed. To speed up convergence and reduce training hours for large models, \cite{kaddour2022stop} proposes LAWA (Latest Weight Averaging), focusing on the middle stages of training. LAWA averages the last checkpoints found at the end of each epoch. Our approach draws inspiration from both  \cite{izmailov2018averaging,kaddour2022stop}, incorporating their intuitions into the federated scenario. By doing so, we aim to improve the performance of federated learning in the presence of heterogeneous data distributions.

\section{\longours}
In this Section, we provide details regarding the objectives of the federated training in cross-device settings (Sec. \ref{subsec:fl}) and introduce the specifics of \ours\ (Sec. \ref{subsec:ours}).

\subsection{Problem formulation}
\label{subsec:fl}
The goal of training in FL is to learn a global model $f(w): \mathcal{X} \rightarrow \mathcal{Y}$, where $\mathcal{X}$ is the input space (\eg, images), $\mathcal{Y}$ the output space (\eg, labels), and $w\in\mathbb{R}^d$ the model parameters. Training proceeds over $T$ communication rounds and is distributed among a set of devices $\mathcal{S}$ (\ie, \emph{clients}), having access to local private datasets $\mathcal{D}_i=\{(x_j,y_j)|\, x_j\in\mathcal{X}, \, y_j\in\mathcal{Y}, \, j\in [N_i], \, i\in\mathcal{S}\}$,  where $N_i=|\mathcal{D}_i|$. We define the overall number of clients $|\mathcal{S}|=:K$ to ease the notation. The global objective is 
\begin{equation}
    \min_{w\in\mathbb{R}^d} F(f_1(w), f_2(w), \ldots, f_K(w)),
\end{equation}
where $F(\cdot)$ is the aggregating function and $f_i\, \forall i\in\mathcal{S}$ is the local objective (\eg, cross-entropy loss). In this work, $F(\cdot)$ is defined by FedAvg as 
\begin{equation}
    \min_{w\in\mathbb{R}^d} \sum_{i\in\mathcal{S}} \frac{N_i}{\sum_{j\in\mathcal{S}} N_j} f_i(w_i),
\end{equation}
where $w_i$ are the locally updated parameters. At each round $t\in[T]$, this minimization problem translates into performing a weighted average of the parameters updated by the subset of selected clients $\mathcal{S}^t$. %
Additionally, \cite{reddi2020adaptive} shows that the FedAvg global update can be generally seen as one step of SGD with unitary learning rate (FedOpt), \ie
\begin{equation}
\label{eq:fedopt}
    w^{t+1}_{\textsc{FedAvg}} =  \sum_{i\in\mathcal{S}^t} \frac{N_i}{N} w_i^t = w^t - \eta_s \sum_{i\in\mathcal{S}^t} \frac{N_i}{N} (w^t - w_i^t),
\end{equation}
where $N = \sum_{i\in\mathcal{S}^t} N_i$ and $\eta_s$ is the server-side learning rate, equal to $1$ in FedAvg. The difference $w^t-w^t_i =: \Delta w^t_i$ defines the $i$-th client's pseudo-gradient, and their average the global pseudo-gradient $\Delta w^t$ at round $t$. The local updates $w_i$ are usually computed using SGD. The server-side update can be also generalized for a generic optimizer as $w^{t+1} = w^t - \textsc{ServerOpt}(w^t, \Delta w^t, \eta_s, t)$, where \textsc{ServerOpt} indicates any optimizer, \eg SGD, Adam \cite{kingma2014adam}, AdaGrad \cite{duchi2011adaptive}.

In realistic settings, local datasets likely follow different distributions, \ie $\mathcal{P}_i \neq \mathcal{P}_j \, \forall i,j\in \mathcal{S}$, resulting in local updates directing towards distinct minima in the typically non-convex loss landscape. This leads to unfavorable behavior, \eg noisy and unstable learning trends, slowed down convergence (Fig. \ref{fig:trends}). In addition, as only a fraction of client $|\mathcal{S}^t| \ll K$ is selected at each round $t$, the resulting model is extremely biased towards the just seen distributions $\mathcal{P}_i\, \forall i\in\mathcal{S}^t$ \cite{caldarola2022improving,luo2021no}, leading to catastrophic forgetting.

\subsection{\textbf{\ours} for Federated Learning}
\label{subsec:ours}
To overcome the instability and bias proper of training in heterogeneous cross-device federated scenarios, in this work, we introduce \textit{\textbf{Wi}ndow-based \textbf{M}odel \textbf{A}veraging} (\ours). Defined a window size of $W$ rounds, at the end of round $t\geq W$, \ours\ averages the last $W$ global models built using FedAvg as:
\begin{equation}
    \label{eq:wima}
    w^{t+1}_{\ours} = w^{t'+W}_{\ours} := \frac{1}{W} \sum_{\tau=t'}^{t'+W-1} w^{\tau+1}_{\textsc{FedAvg}},
\end{equation}
where $t'=t+1-W$ is the first round comprised in the window frame.  %
The rationale behind this approach is to enhance robustness of the global model towards distribution shifts across rounds and diminish bias towards the last-seen clients by averaging models that are still experiencing significant changes. By considering the last $W$ rounds, we retain sufficient history to stabilize the model without hindering the training process, as observed with SWA. %

\subsubsection{Unveiling the window contents}
\label{subsec:decay}
We now try to answer the question \textit{``What information is stored inside the window?"} To do so, we reformulate Eq.~\ref{eq:wima} using the updates provided in Eq.~\ref{eq:fedopt}:
\begin{align}
    w^{t'+W}_\ours &= \frac{1}{W} \sum_{\tau=t'}^{t'+W-1} w^{\tau+1}_\textsc{FedAvg} \tag{Eq.~\ref{eq:wima}}\\
    &= \frac{1}{W} \sum_{\tau=t'}^{t'+W-1} \sum_{i\in\mathcal{S}^\tau}\frac{N_i}{N} w_i^\tau \tag{FedAvg in Eq.~\ref{eq:fedopt}}\\
    &= \frac{1}{W} \sum_{\tau=t'}^{t'+W-1} \big(w^\tau - \eta_s \sum_{i\in\mathcal{S}^\tau}\frac{N_i}{N} (w^\tau - w_i^\tau) \big).
\end{align}

By unraveling the summation over the last $W$ rounds and writing each update using Eq.~\ref{eq:fedopt}, we find out that the \ours\ model's update is equivalent to
\begin{equation}
\label{eq:lr_decay}
    w^{t'} - \eta_s \sum_{\tau=t'}^{t'+W-1} \frac{t'+W-\tau}{W} \sum_{i\in\mathcal{S}^\tau} \frac{N_i}{N} (w^\tau - w_i^\tau ),
\end{equation}
or more in general
\begin{equation}
\label{eq:general_decay}
    w^{t'} - \sum_{\tau=t'}^{t'+W-1} \frac{t'+W-\tau}{W} \textsc{ServerOpt}(w^\tau, \Delta w^\tau, \eta_s, \tau).
\end{equation}
The term $\nicefrac{t'+W-\tau}{W}$ tends to $1$ when $\tau=t'$, \ie at the beginning of the queue, and to $\nicefrac{1}{W}$ when $\tau=t'+W-1$, \ie in the last round. Thus, Eq.~\ref{eq:general_decay} can be interpreted as $W-1$ SGD steps starting from the initial model $w^{t'}$  with a \textbf{\textit{learning rate decay}} that depends on the position in the queue, given by $\nicefrac{t'+W-\tau}{W}$. Indeed, \ours\ assigns higher significance to previous updates, as they are perceived as more stable, while also integrating new knowledge at a rate proportional to the window size $W$. This sets it apart from methods like momentum, which prioritize more recent updates. Additional details can be found in Appendix~\ref{appendix}.
\section{Experiments}
In this Section, we provide numerical results on the application of \ours\ to different heterogeneous federated scenarios. Sec.~\ref{subsec:impl} informs on datasets used, model architectures, and training details. Final results and comparison with state-of-the-art approaches can be found in Sec.~\ref{subsec:exps}, while Sec.~\ref{subsec:ablation} studies \ours\ more in depth.

\subsection{Implementation details}
\label{subsec:impl}
Here we provide a detailed description of the experimental settings. Large-scale experiments were performed using an NVIDIA DGX A100, while the others run on one NVIDIA GeForce GTX 1070. The code was built starting from the FedJAX framework \cite{fedjax2021}. All runs are averaged over $3$ seeds.

\subsubsection{Datasets}
\setlength\tabcolsep{3.2pt}

\begin{table}[t]
\small
\caption{Datasets statistics. Clf-$X$ indicates the classification task over $X$ classes, while NCP stands for Next Character Prediction.}
    \begin{center}
    \begin{tabular}{lcccc}
    \toprule
         {\textbf{Dataset}} & {\textbf{Distribution}} & {\textbf{Task}} & {\textbf{Clients}} & \textbf{Imbalance}\\
    \hline
    \textsc{Cifar10} & $\alpha=0, 0.05$ & Clf-$10$ & $100$ & \ding{55}\\
    \multirow{2}{*}{\textsc{Cifar100}} & $\alpha=0, 0.5$ & Clf-$100$ & $100$ & \ding{55}\\ %
    & PAM & Clf-$100$ & $500$ & \ding{55}\\
    \textsc{Femnist} & NIID & Clf-$62$ & $3,400$ & \ding{51}\\
    \textsc{GldV2} & NIID & Clf-2,028 & 1,262 & \ding{51}\\
    \textsc{Shakespeare} & NIID & NCP & $715$ & \ding{51}\\
    \bottomrule
    \end{tabular}
    \end{center}
    \label{tab:datasets}
\end{table}
Table \ref{tab:datasets} summarizes the information on the used datasets, chosen among common FL benchmarks. As for vision tasks, we focus on classification and use the federated $\textsc{Cifar10}$, $\textsc{Cifar100}$ \cite{Krizhevsky09learningmultiple} and $\textsc{Femnist}$ \cite{caldas2018leaf}. We introduce large-scale experiments on \textsc{Landmarks-User-160k} \cite{hsu2020federated}, the federated version of \textsc{Google Landmarks v2} \cite{weyand2020google}, which we will refer to as \textsc{Gldv2} for short. To further prove the wide applicability of our method, we additionally test it on \textsc{Shakespeare} \cite{caldas2018leaf} for the next character prediction task. The $\alpha$ value in Table \ref{tab:datasets} refers to the parameter of the latent Dirichlet's distribution applied to the labels, as proposed by \cite{hsu2019measuring}. A smaller value of $\alpha$ identifies a more skewed setting, with $\alpha=0$ being its extreme scenario in which each client only sees one class. \textsc{Cifar100/PAM} leverages the Pachinko Allocation
Method \cite{li2006pachinko} instead. More details can be found in \cite{reddi2020adaptive}. \textsc{Femnist} is split according to the writer's information, while clients in \textsc{Shakespeare} correspond to characters in Shakespearean plays and each user is the author of the picture in \textsc{GldV2}. Images are pre-processed using standard data augmentation techniques, \eg random crop, horizontal flip.

\subsubsection{Models} We use a ResNet20 \cite{he2016residual} on all the distributions of the \textsc{Cifar} datasets, substituting Batch Normalization with Group Normalization layers, as suggested by \cite{hsieh2020non}. For \textsc{Femnist} and \textsc{Shakespeare} we use the architectures proposed in FedJAX, a $2$-layer Convolutional Neural Network and an LSTM~\cite{hochreiter1997long} network respectively, following \cite{mcmahan2017communication,caldas2018leaf}. As done in \cite{hsu2020federated,caldarola2022improving}, we train MobileNetV2 \cite{sandler2018mobilenetv2} pre-trained on ImageNet \cite{deng2009imagenet} for \textsc{GldV2}, replacing Batch Normalization layers with Group Normalization ones. 

\subsubsection{Training details} In all cases, on the server side, we use the standard FedAvg with $\eta_s=1$ unless otherwise specified and momentum $0$, and clients locally train with SGD. Experiments on the Dirichlet's \textsc{Cifar} datasets are run for $10k$ rounds, selecting $10$ clients at each round, \ie with $10\%$ participation rate. On the client side, we select learning rate $0.1$ from $\{0.1, 0.01\}$, momentum $0$ from $\{0,0.9\}$, weight decay $0$ unless otherwise specified, batch size $100$ among $\{32,64,100,128\}$, and train for $1$ local epoch chosen from $\{1,2,4\}$. For \textsc{Cifar100/PAM} we train for $10k$ rounds with $20\%$ participation rate, using learning rate $0.05$, weight decay $4e$-$4$, batch size $20$, server-side momentum $0.9$ from \cite{caldarola2022improving}. For \textsc{Femnist} we use client learning rate $0.1$ from $\{0.1,0.01,0.001\}$, momentum $0$ from $\{0,0.9\}$, weight decay $0$, batch size $10$ from $\{10,20,32\}$. We train for $1,500$ rounds with $10$ clients per round ($\approx 0.3\%$ participation rate), performing $1$ local epoch each. For \textsc{GldV2}, we follow the setup of \cite{caldarola2022improving} except for the batch size equal to $50$ and train the model for $3k$ rounds with $10$ clients selected at the time. In \textsc{Shakespeare}, local learning rate is $1$, momentum $0$, weight decay $0$, batch size $4$, $1$ epoch from \cite{karimireddy2020mime}. Training is spanned over $1,500$ rounds with 10 clients per round ($\approx 1.4\%$ participation rate). The \ours\ parameter $W$ is set to $100$ for all settings except for \textsc{GldV2}, where $W=370$ (see Sec.~\ref{subsec:ablation} for additional analyses). For all datasets, the reported final results are averaged over the last $100$ rounds for increased robustness \cite{caldarola2022improving}. %

\begin{figure}[t]
    \centering
    \includegraphics[width=.9\linewidth]{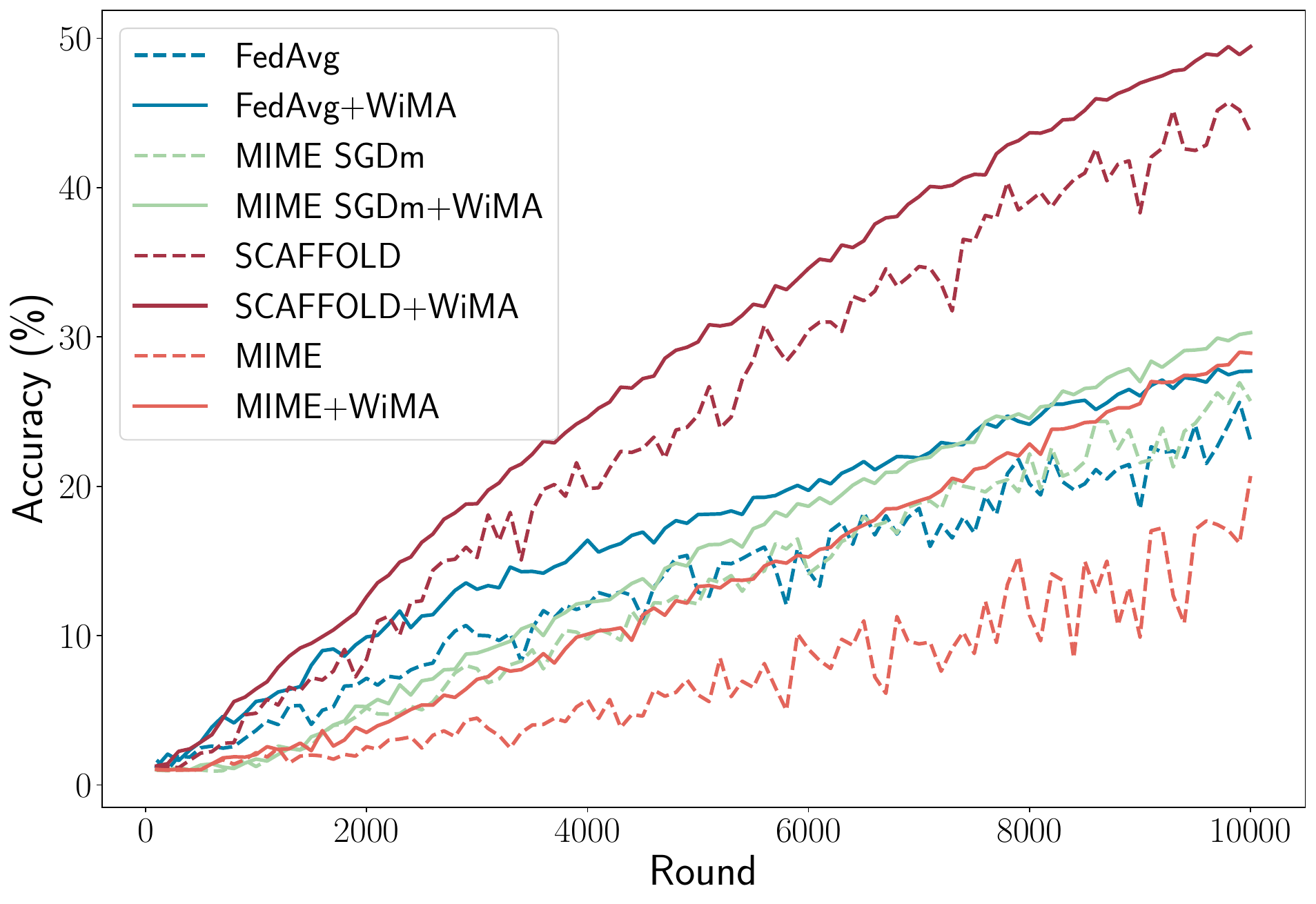}
    \caption{Accuracy trends of different SOTA algorithms on \cifar\ $\alpha=0$ across rounds, with and without \ours\ (dashed lines). %
    The application of \ours\ results in smoother and more stable trends, leading to enhanced robustness and improved performance.
    Best seen in colors.}
    \label{fig:trends}
\end{figure}

\subsubsection{SOTA algorithms details} We provide here details on the tuning intervals for the state-of-the-art (SOTA) algorithms used for comparison. %
We apply \ours\ on top of methods proposed for addressing statistical heterogeneity in FL. Looking at momentum-based approaches, we select FedAvgM \cite{hsu2019measuring} (server-side momentum $\beta=0.9$, $\eta_s\in\{0.1, 1\}$), MIME SGD ($\eta_s\in\{0.1, 1\}$) and SGDm \cite{karimireddy2020mime}, \ie with momentum $0.9$, MIMELite SGDm \cite{karimireddy2020mime} ($\eta_s\in\{0.1, 1\}$, momentum $0.9$), FedCM \cite{xu2021fedcm} ($\alpha_{\textsc{cm}}\in\{0.05, 0.1, 0.5\}$) and FedACG \cite{kim2022communication} ($\beta_\textsc{acg}\in\{0.01, 0.001\}$, $\lambda_\textsc{acg}\in\{0.8,0.85,0.9\}$). We additionally test SCAFFOLD \cite{karimireddy2020scaffold}, FedProx \cite{li2020federated_fedprox} ($\mu_\textsc{prox}\in\{0.1, 0.01, 0.001\}$), FedDyn \cite{acar2021federated} ($\alpha_\textsc{dyn}\in\{0.01, 0.001\}$) and AdaBest \cite{varno2022adabest} ($\mu_\textsc{adabest}\in\{0.01,0.02\}$, $\beta_\textsc{adabest}\in\{0.5, 0.6, 0.7, 0.8, 0.9, 0.95\}$) to reduce the client drift. Lastly, we compare \ours\ with \textsc{SWA} applied from $75\%$ of training onwards, for which we use $c\in\{10,20\}$ and second learning rate equal to $\eta \cdot 10^{-2}$, following \cite{caldarola2022improving}.

\subsection{Results}
\label{subsec:exps}

\paragraph{Reducing noise and increasing stability with \ours.} Thanks to the window-based average of global models, \ours\ mitigates the negative impact of statistical heterogeneity inherent in cross-device federated settings. As shown in Fig.~\ref{fig:trends}, \ours\ effectively smooths the learning trends, resulting in enhanced robustness and reduced instability. Notably, these benefits are observed across all performance levels, with improvements evident in low-performing approaches (\eg, MIME in Fig.~\ref{fig:trends}) as well as the best-performing ones (\eg, SCAFFOLD).

\begin{figure}[t]
    \centering
    \includegraphics[width=0.9\linewidth]{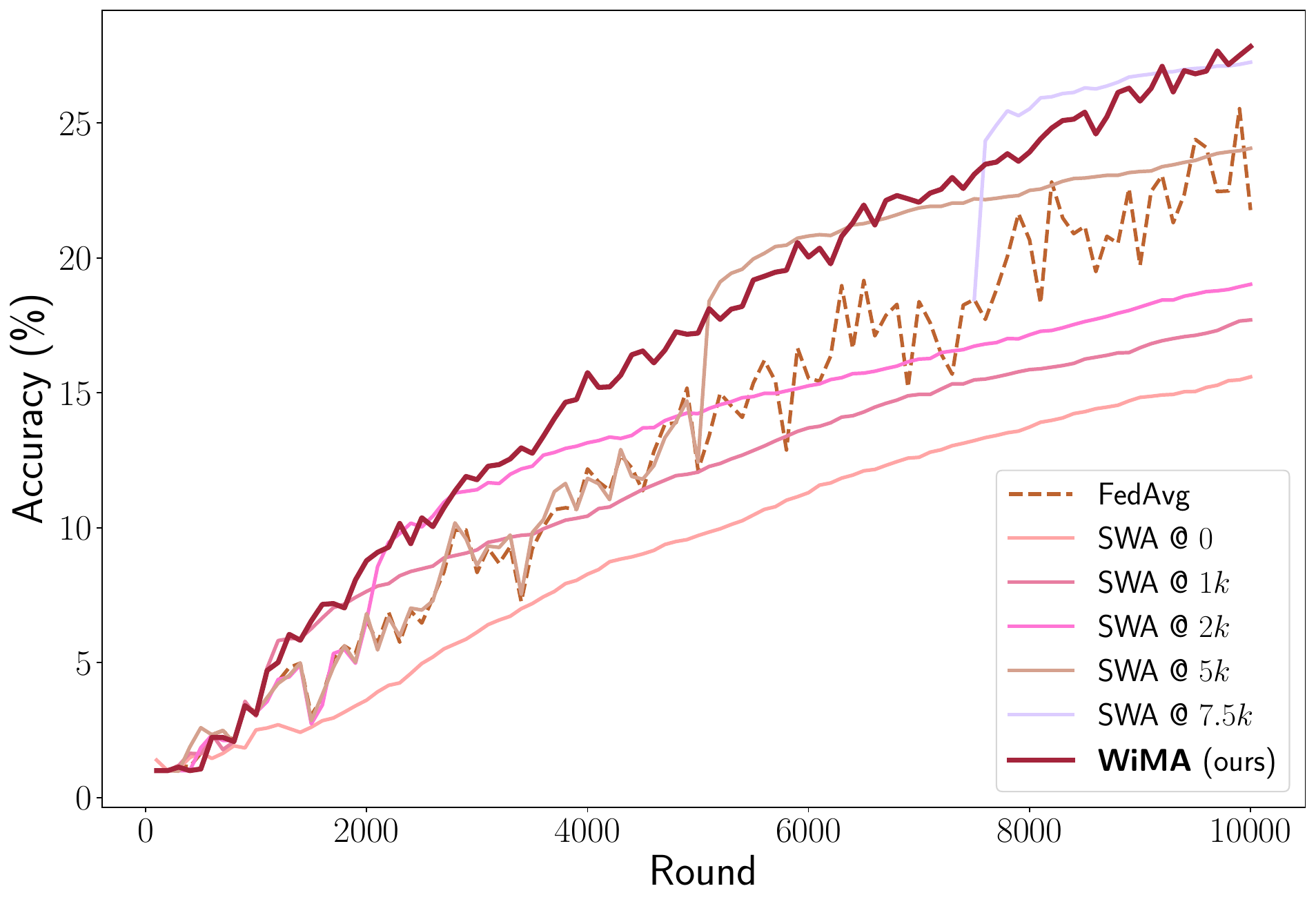}
    \caption{Accuracy trends of \ours\ and \textsc{SWA} starting at different rounds on \cifar\ with $\alpha=0$, using FedAvg as reference. \ours\ has a stable trend from the beginning, leading to final better performances, while \textsc{SWA} suffers from early initialization. Best seen in colors.}
    \label{fig:swa}
\end{figure}

\paragraph{The effective combination of \ours\ with SOTA.}
\setlength\tabcolsep{2.8pt}

\begin{table*}[]
\scriptsize
\caption{\ours\ combined with state-of-the-art FL algorithms. For each configuration, the first column reports the accuracy ($\%$) reached by each standalone method; in the second column, the performance achieved when adding \ours. Between brackets the improvements introduced by \ours, underlined the best ones in each dataset. For simplicity, we only report gains in improvements $\geq 1.5$. Best overall accuracy in bold.}
    \begin{center}
\begin{tabular}{lcc|cc|cc|cc|cc|cc|cc}
    \toprule
        \multirow{3}{*}{\textbf{Algorithm}} & \multicolumn{4}{c}{\textsc{\textbf{Cifar10}}} & \multicolumn{6}{c}{\textsc{\textbf{Cifar100}}} & \multicolumn{2}{c}{\textsc{\multirow{2}{*}{\textbf{Femnist}}}} & \multicolumn{2}{c}{\textsc{\multirow{2}{*}{\textbf{Shakespeare}}}}\\
        \cmidrule(l){2-5}\cmidrule(l){6-11}
        &  \multicolumn{2}{c}{$\alpha=0$} & \multicolumn{2}{c}{$\alpha=0.05$} & \multicolumn{2}{c}{$\alpha=0$} & \multicolumn{2}{c}{$\alpha=0.5$} & \multicolumn{2}{c}{\textsc{PAM}} &  & \\
        &  & w/ \ours &   & w/ \ours &   & w/ \ours &   & w/ \ours &   & w/ \ours &   & w/ \ours &   & w/ \ours\\
        \hline
        FedAvg & 64.37&69.95 \tiny{(\textcolor{ForestGreen}{$\uparrow 5.3$})}& 68.50 & 72.69 \tiny{(\textcolor{ForestGreen}{$\uparrow 4.2$})}& 23.00 & 27.91 \tiny{(\textcolor{ForestGreen}{$\uparrow 4.9$})} & 31.21 & 34.45\tiny{(\textcolor{ForestGreen}{$\uparrow 3.2$})} & 47.41 & 48.53 
        & 83.59 & \underline{85.06} \tiny{(\textcolor{ForestGreen}{$\uparrow$ \boldmath$1.5$})} & \textbf{56.86} & 57.74 \\
        FedAvgM & 73.32& 75.72 \tiny{(\textcolor{ForestGreen}{$\uparrow 2.4$})} & 73.10 & 75.30 \tiny{(\textcolor{ForestGreen}{$\uparrow 2.2$})} & 24.27 & 28.77 \tiny{(\textcolor{ForestGreen}{$\uparrow 4.5$})} & 31.78 & 33.97 \tiny{(\textcolor{ForestGreen}{$\uparrow 2.2$})} & 55.96 & 61.63 \tiny{(\textcolor{ForestGreen}{$\uparrow 5.7$})} & 85.00 & 85.26 
        &  56.91 & 57.57 \\
        MIME SGD & 74.92 & 80.65 \tiny{(\textcolor{ForestGreen}{$\uparrow 5.7$})} & 78.82 &  82.81 \tiny{(\textcolor{ForestGreen}{$\uparrow 4.0$})} & 17.55 & \underline{29.05} \tiny{(\textcolor{ForestGreen}{\textbf{$\uparrow$ \boldmath$11.5$}})} & 27.30 & \underline{40.37} \tiny{(\textcolor{ForestGreen}{$\uparrow$ \boldmath$ 13.1$})} & 54.33 & 57.44 \tiny{(\textcolor{ForestGreen}{$\uparrow 3.1$})} & 85.37 & 86.40 
        & 56.06 & 57.43 \\
        MIME SGDm & 74.58& 76.20 \tiny{(\textcolor{ForestGreen}{$\uparrow 1.6$})} &78.39 & 80.38 \tiny{(\textcolor{ForestGreen}{$\uparrow 2.0$})} & 25.78 & 30.11 \tiny{(\textcolor{ForestGreen}{$\uparrow 4.3$})} & 38.42 & 43.08 \tiny{(\textcolor{ForestGreen}{$\uparrow 4.7$})} & 54.62 & 57.28 \tiny{(\textcolor{ForestGreen}{$\uparrow 2.7$})} & 86.67 &	87.40 
        & 54.00 & 54.68 \\
        MIMELite & 64.42 & 67.78 \tiny{(\textcolor{ForestGreen}{$\uparrow 3.4$})} & 68.27 & 71.21 \tiny{(\textcolor{ForestGreen}{$\uparrow 2.9$})} & 20.00 & 24.69 \tiny{(\textcolor{ForestGreen}{$\uparrow 4.7$})} & 35.56 & 39.15 \tiny{(\textcolor{ForestGreen}{$\uparrow 3.6$})} & 53.97 & \underline{60.34} \tiny{(\textcolor{ForestGreen}{$\uparrow$ \boldmath$ 6.4$})} & \textbf{86.82} & \textbf{87.51 }%
        & 52.45 & 53.01 \\
        FedCM & 78.83 & 81.73 \tiny{(\textcolor{ForestGreen}{$\uparrow 2.9$})} & 73.94 & \underline{80.28} \tiny{(\textcolor{ForestGreen}{$\uparrow$ \boldmath$6.3$})} &19.62 & 25.29 \tiny{(\textcolor{ForestGreen}{$\uparrow 5.7$})} & 36.12 & 40.10 \tiny{(\textcolor{ForestGreen}{$\uparrow 4.0$})} & 53.16 & 54.12 
        & 83.88 & 84.90 
        &  38.90	&39.29\\
        FedACG & 55.27 & 60.09 \tiny{(\textcolor{ForestGreen}{$\uparrow 4.8$})}& 63.20 & 66.35 \tiny{(\textcolor{ForestGreen}{$\uparrow 3.2$})} & 20.09 & 23.55 \tiny{(\textcolor{ForestGreen}{$\uparrow 3.5$})} & 29.74 & 32.46 \tiny{(\textcolor{ForestGreen}{$\uparrow 2.7$})} & 58.88 & 61.38 \tiny{(\textcolor{ForestGreen}{$\uparrow 2.5$})} & 85.73 & 86.14 
        & 56.79&	58.03 \\
        FedProx & 64.25 & 69.90 \tiny{(\textcolor{ForestGreen}{$\uparrow 6.7$})}& 67.82 & 71.90 \tiny{(\textcolor{ForestGreen}{$\uparrow 4.1$})} & 22.59 & 27.58 \tiny{(\textcolor{ForestGreen}{$\uparrow 5.0$})} & 30.70 & 33.68 \tiny{(\textcolor{ForestGreen}{$\uparrow 3.0$})} & 55.91 & 62.25 \tiny{(\textcolor{ForestGreen}{$\uparrow 6.3$})} & 84.50 & 85.21 
        & 55.92&	56.71 \\
        SCAFFOLD & \textbf{81.45} & \textbf{83.96} \tiny{(\textcolor{ForestGreen}{$\uparrow 2.5$})}& \textbf{83.24} & \textbf{85.17} \tiny{(\textcolor{ForestGreen}{$\uparrow 1.9$})} & \textbf{45.65} & \textbf{49.77} \tiny{(\textcolor{ForestGreen}{$\uparrow 4.1$})} & \textbf{50.93} & \textbf{53.75} \tiny{(\textcolor{ForestGreen}{$\uparrow 2.8$})} & 56.09 & 57.64 \tiny{(\textcolor{ForestGreen}{$\uparrow 1.6$})} & 85.87	& 86.61 & 
        56.68&	57.48\\
        FedDyn & N/A& N/A& N/A& N/A & 5.88 & 8.48 \tiny{(\textcolor{ForestGreen}{$\uparrow 2.6$})} & 20.88 & 24.54 \tiny{(\textcolor{ForestGreen}{$\uparrow 3.7$})} & \textbf{57.42} & \textbf{63.00} \tiny{(\textcolor{ForestGreen}{$\uparrow 5.6$})} & N/A & N/A & 54.54	& 55.09 \\
        AdaBest & 66.05 & \underline{73.95} \tiny{(\textcolor{ForestGreen}{$\uparrow$ \boldmath$7.9$})} & 71.54& 77.42 \tiny{(\textcolor{ForestGreen}{$\uparrow 5.9$})} & 24.92 & 31.41 \tiny{(\textcolor{ForestGreen}{$\uparrow 6.5$})} & 37.45 & 43.81 \tiny{(\textcolor{ForestGreen}{$\uparrow 6.4$})}& 54.98 & 57.57 \tiny{(\textcolor{ForestGreen}{$\uparrow 2.6$})} &  84.95 & 86.02 & 56.60&	\underline{\textbf{58.12}} \tiny{(\textcolor{ForestGreen}{$\uparrow$ \boldmath$1.5$})}\\
    \bottomrule
    \end{tabular}
    \end{center}
    \label{tab:sota}
\end{table*} Table \ref{tab:sota} presents the results achieved by combining \ours\ with SOTA federated algorithms designed to handle statistical heterogeneity. Looking at standalone algorithms (\ie, w/o \ours), SCAFFOLD achieves the best performances overall. FedDyn is not able to converge in the most heterogeneous settings, as already shown by \cite{varno2022adabest,caldarola2022improving}. Notably,  \ours\ enables each method to achieve better final accuracy, showcasing substantial improvements compared to the algorithm without \ours. The most significant gains are observed on the more challenging \textsc{Cifar} datasets. In particular, \ours\ proves especially beneficial for the worst-performing methods, increasing the final accuracy by over $11$ points for MIME SGD on both $\alpha$ values in \textsc{Cifar100}. On the other hand, using the aggregation proposed by \ours\ is effective even on the overall best-performing SCAFFOLD, or on the less challenging \textsc{Femnist} and \textsc{Shakespeare} datasets. Thus, all methods and settings are positively affected by the increased robustness and stability introduced by \ours.

\paragraph{\ours\ in large-scale classification.} In Table \ref{tab:gldv2}, we introduce the results obtained when using \ours\ for large-scale classification on \textsc{GldV2}. Without redundancy and loss of generality, we present the performance of \ours\ when integrated with both the standard FedAvg and the best-performing SCAFFOLD. Even in this more complex vision scenario, \ours\ achieves large gains in accuracy. 

\setlength\tabcolsep{3pt}

\begin{table}[h]
\caption{Large-scale experiments. Results in test accuracy (\%) on \textsc{GldV2}. Best result in bold.}
    \small
    \begin{center}
    \begin{tabular}{lcc}
    \toprule
         \textbf{Algorithm} & \textbf{w/o \ours} & \textbf{w/ \ours}\\ \hline
         FedAvg & 58.17 & 63.05\\
         SCAFFOLD & 62.32 & \textbf{68.30}\\
    \bottomrule
    \end{tabular}
    \end{center}
    \label{tab:gldv2}
\end{table}

\paragraph{\ours\ vs \textsc{SWA}.} Fig. \ref{fig:swa} compares the accuracy trends of \ours\ and \textsc{SWA} starting at different rounds. We note that \textsc{SWA} suffers from early initialization, leading to saturation and worse performances than FedAvg, reaching a final accuracy comparable to our method only if close to the last rounds. Differently, thanks to the windowed view of the global models across rounds, \ours\ can be applied from the beginning of training, and presents a constantly stable and better trend than \textsc{SWA}.

\begin{figure}
    \centering
    \subfloat[\centering $\alpha=0$]{{\includegraphics[width=.5\linewidth]{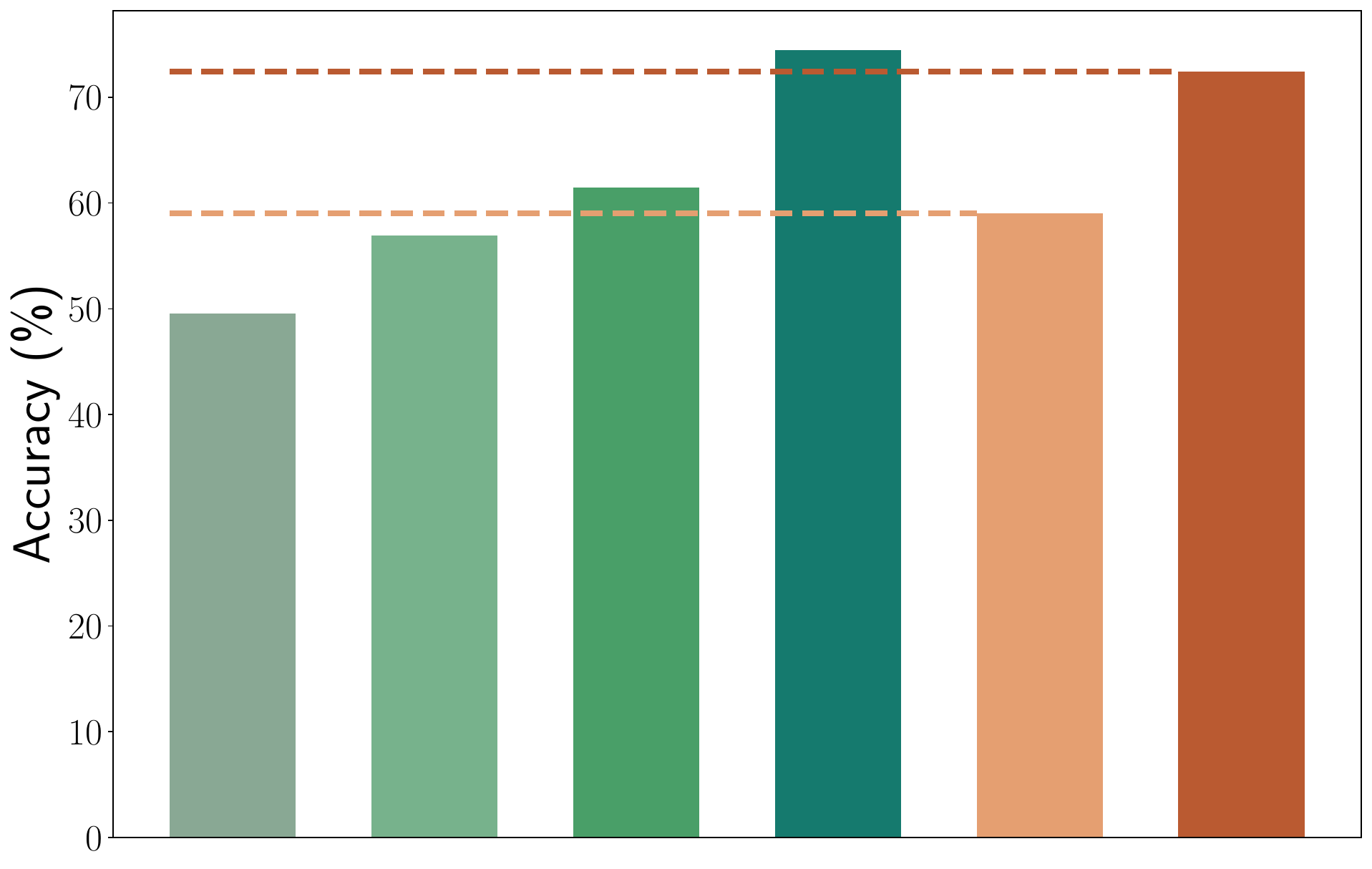} }}%
    \subfloat[\centering $\alpha=0.05$]{{\includegraphics[width=.5\linewidth]{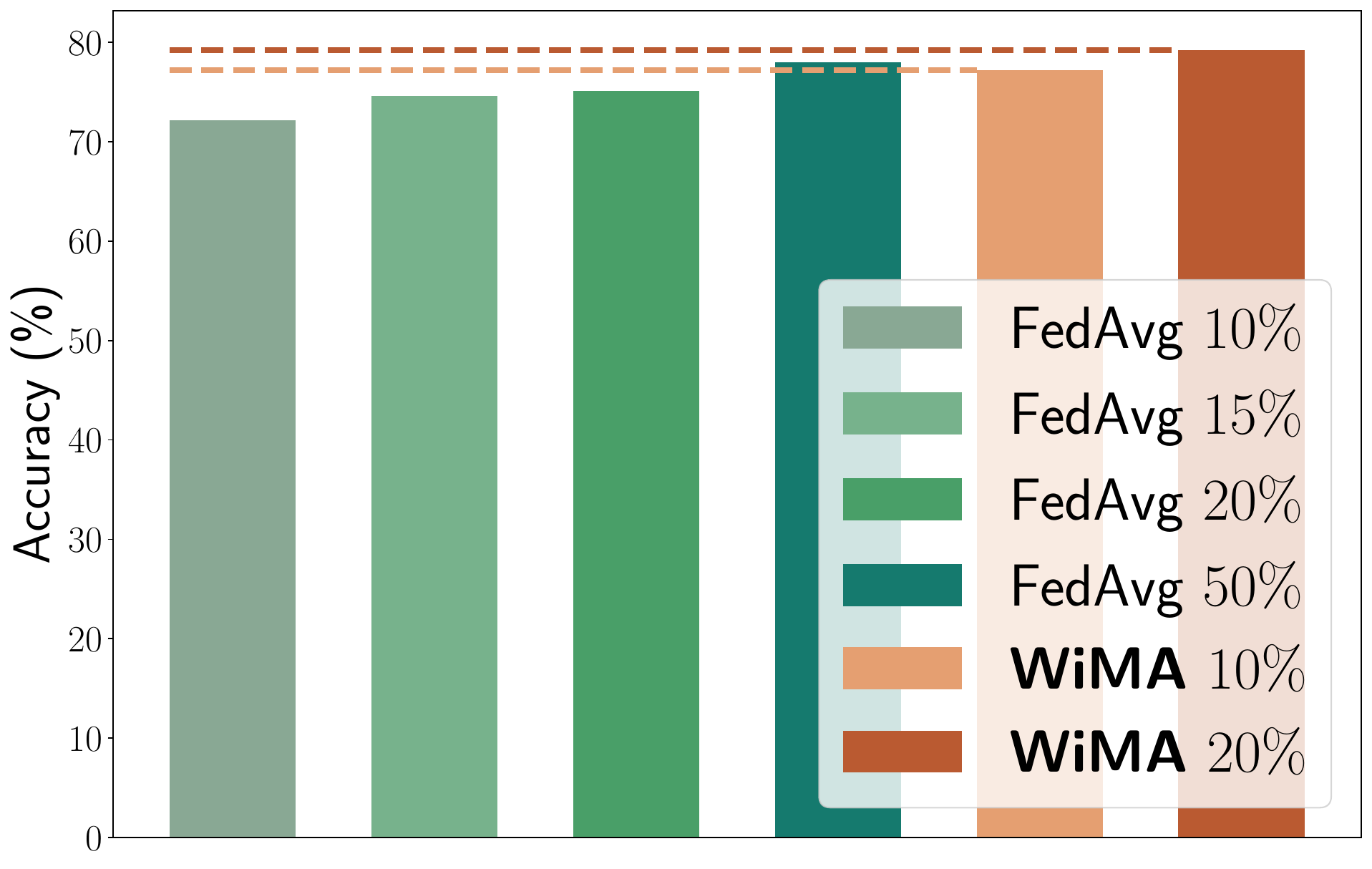} }}%
    \caption{\ours\ performances compared with varying client participation rates at each round on \cifarten\ using FedAvg. \textbf{a)} \ours\ achieves higher accuracy with $10\%$ participation compared to FedAvg with $1.5$ times the number of devices per round. \ours\ with $20$ clients perform similarly to FedAvg with half the clients. 
    \textbf{b)} \ours\ with $10\%$ rate performs almost on par with FedAvg w/o \ours\ selecting $50\%$ of the devices.}%
    \label{fig:rates}
\end{figure}

\paragraph{\ours\ allows less client participation.} In Fig.~\ref{fig:rates}, we observe that the enhanced generalization capability achieved with \ours\ allows narrowing the gap with runs involving higher client participation rates. Specifically, we compare FedAvg training with and without our method, using varying numbers of clients selected at each round on \cifarten. Experiments are run using batch size $20$ to account for more local iterations, highlighting the client drift. \ours\ enables the model with a $10\%$ participation rate to attain a final accuracy that is \textit{at least} comparable to the run involving $1.5$ times the number of devices with $\alpha=0$ and twice that number with $\alpha=0.05$. When $20\%$ of clients are involved instead, \ours\ reaches performances comparable ($\alpha=0$) or better ($\alpha=0.05$) than FedAvg involving half the devices ($50\%$ rate). This result holds significant importance in cross-device settings, where devices are often unavailable due to factors such as limited battery life, network connectivity issues, and communication overload \cite{kairouz2021advances}. The ability to achieve improved results with fewer clients involved aligns favorably with real-life requirements, making it a valuable contribution. %

\subsection{Ablation study}
\label{subsec:ablation}
\paragraph{Studying the window size.} The dimension $W$ of the window used by \ours\ plays a crucial role in achieving a trade-off between retaining useful historical information and avoiding excessively old data. In Table \ref{tab:window} we compare the accuracy reached with varying values of $W$ on the heterogeneous \textsc{Cifar}s. Smaller $W$ values lead to lower performance as the \ours\ model fails to capture sufficient information from the underlying distribution, while excessively large $W$ values slow down training by relying on outdated updates. %
The optimal results are obtained with $W = 100$ in both cases.
\setlength\tabcolsep{6pt}
\begin{table}[h]
\caption{\ours\ accuracy (\%) with varying $W$ on the \textsc{Cifar} datasets with $\alpha=0$. Best results in bold.}
\small
\begin{center}
    \begin{tabular}{ccc}
    \toprule
        \multirow{2}{*}{\textbf{Window size $W$}} & \multicolumn{2}{c}{\textbf{\ours\ Accuracy} (\%)} \\
        \cmidrule{2-3}
        & \textsc{Cifar10} & \textsc{Cifar100}\\
        \hline
        $5$ & $67.25$ & $22.71$\\
        $10$ & $69.12$ & $25.70$\\
        $50$ & $69.79$& $22.75$\\
        $100$ & \boldmath$69.94$& \boldmath$27.91$\\
        $200$ & $68.74$& $27.72$\\
        \bottomrule
    \end{tabular}
    \end{center}
    \label{tab:window}
\end{table}

\paragraph{\ours\ outputs better features.} We now wonder where \ours\ helps the model the most. In particular, with the goal of understanding which part of the architecture our method affects the most, we evaluate its performances when acting only on the feature extractor, or the classifier, \ie the last linear layer of the model. To allow for more client-side finetuning, we use a batch size of $20$. The analyses reported in Table \ref{tab:feat} demonstrate that \ours\ is mainly acting on the feature extractor. Thanks to the more robust and less biased output features, the classifier is consequently able to give more accurate predictions. 
\setlength\tabcolsep{3pt}

\begin{table}[h]
\caption{Accuracy (\%) reached when applying \ours\ only on the classifier (\textit{clf}), the feature extractor (\textit{feat. extr.}) or all the model parameters (\textit{all}) as reference. }
    \small
    \begin{center}
    \begin{tabular}{cccc|c}
    \toprule
         \textbf{Dataset} & \boldmath$\alpha$ & \textbf{\ours\ \textit{clf}} & \textbf{\ours\ \textit{feat. extr.}} & \textbf{\ours\ \textit{all}} \\
         \hline
         \multirow{2}{*}{\textsc{Cifar10}}& $0$ & $47.76$ & $59.03$ & $59.53$\\
         & $0.05$ & $71.01$ & $76.72$ & $78.87$\\
         \hline
         \multirow{2}{*}{\textsc{Cifar100}} & $0$ & $25.13$ & $27.10$ & $27.91$\\
         & $0.5$ & $36.12$ & $36.29$ & $37.88$\\
    \bottomrule
    \end{tabular}
    \end{center}
    \label{tab:feat}
\end{table}
\section{Conclusions}
In this work, we proposed \textit{\longours}\ (\ours) to address the negative impacts of statistical heterogeneity in federated learning scenarios. In particular, our goal is to reduce the noise and instability proper of learning trends of models trained in non-\textit{i.i.d}. federated settings. To addresses these issues, \ours\ averages the last $W$ global models built using any server-side optimizer at each round. Thanks to the windowed view of the rounds, we keep sufficient history to stabilize the model without hindering the training process. \ours\ can be easily combined with most of the existing state-of-the-art algorithms, significantly improving the performance of each method and leading to smoother and more stable trends. We showed that \ours\ mainly affects the backbone of the network, producing better output features and consequently enabling the classifier in giving more accurate predictions. Lastly, \ours\ helps narrowing the gap with runs using higher client participation rates, a favorable result for realistic federated settings.
\\

\noindent{\large{\textbf{Acknowledgments}}}
This study was carried out within the FAIR - Future Artificial Intelligence Research and received funding from the European Union Next-GenerationEU (PIANO NAZIONALE DI RIPRESA E RESILIENZA (PNRR) – MISSIONE 4 COMPONENTE 2, INVESTIMENTO 1.3 – D.D. 1555 11/10/2022, PE00000013). This manuscript reflects only the authors’ views and opinions, neither the European Union nor the European Commission can be considered responsible for them. Large scale experiments were run using the CINECA infrastructure. We thank the anonymous reviewers for their valuable feedback.

{\small
\bibliographystyle{ieee_fullname}
\bibliography{biblio}
}

\clearpage
\onecolumn
\appendix
\section{Appendix}
\label{appendix}
In Sec. \ref{subsec:decay}, we describe how the \ours\ update is equivalent to updating the first model comprised in the window frame $w^{t'}$ with various SGD steps, using a learning rate decay dependent on the position in the queue, given by $\nicefrac{t'+w-\tau}{W}$ (Eq.~\ref{eq:lr_decay},\ref{eq:general_decay}). We describe here the steps to reach this conclusion.

We recall that 
\begin{align}
    w^{t'+W}_\ours &= \frac{1}{W} \sum_{\tau=t'}^{t'+W-1} w^{\tau+1}_\textsc{FedAvg} \tag{Eq.~\ref{eq:wima}}\\
    &= \frac{1}{W} \sum_{\tau=t'}^{t'+W-1} \sum_{i\in\mathcal{S}^\tau}\frac{N_i}{N} w_i^\tau \tag{FedAvg in Eq.~\ref{eq:fedopt}}\\
    &= \frac{1}{W} \sum_{\tau=t'}^{t'+W-1} \big(w^\tau - \eta_s \sum_{i\in\mathcal{S}^\tau}\frac{N_i}{N} (w^\tau - w_i^\tau) \big), \tag{FedOpt in Eq.~\ref{eq:fedopt}}
\end{align}
where $w^{\tau+1}_\textsc{FedAvg}$ is the new global model built with FedAvg at the end of round $\tau$, $W$ the window size, $t'$ the first round comprised in window frame, $w_i$ the local update of client $i$, $\mathcal{S}^t$ the subset of clients selected at round $t$, $\eta_s$ the server learning rate.

For simplicity, we first assume all clients have access to the same number of images, \ie $\frac{N_i}{N} = \frac{1}{|\mathcal{S}^t|}$. Since the same number of clients is selected at each round, $\frac{1}{|\mathcal{S}^t|} = \frac{1}{|\mathcal{S}|^{t-1}}$.

First, we recursively rewrite $w^\tau$ following Eq.~\ref{eq:fedopt} as
\begin{align}
    w_{\text{\ours}}^{t'+W} &= \frac{1}{W} \sum_{\tau=t'}^{t'+W-1} \Big(w^{\tau} - \frac{1}{|\mathcal{S}^{\tau}|} \sum_{i\in\mathcal{S}^{\tau}}  (w^{\tau} - w_i^{\tau})\Big) \\
    &=\frac{1}{W} \sum_{\tau=t'}^{t'+W-1} \Big( w^\tau - \frac{1}{|\mathcal{S}^{\tau}|} \sum_{i\in\mathcal{S}^{\tau}} \big(\underbrace{w^{\tau-1} - \frac{1}{|\mathcal{S}^{\tau-1}|} \sum_{j\in\mathcal{S}^{\tau-1}} (w^{\tau-1} - w_j^{\tau-1})}_{w^\tau} - w^\tau_i \big)\Big) \\
    &\stackrel{|\mathcal{S}^{\tau-1}| = |\mathcal{S}^{\tau}|}{=} 
     \frac{1}{W} \sum_{\tau=t'}^{t'+W-1} \Big( w^\tau - \frac{1}{|\mathcal{S}^{\tau}|} \sum_{i\in\mathcal{S}^{\tau}} \big(w^{\tau-1} - \frac{1}{|\mathcal{S}^{\tau}|} \sum_{j\in\mathcal{S}^{\tau-1}} (\underbrace{w^{\tau-2} - \frac{1}{|\mathcal{S}^{\tau}|} \sum_{l\in\mathcal{S}^{\tau-2}} (w^{\tau-2} - w_l^{\tau-2})}_{w^{\tau-1}} +\\
     &- w_j^{\tau-1}) -w_i^\tau \big)\Big) \\
     &= \ldots = \frac{1}{W} \sum_{\tau=t'}^{t'+W-1} \Big( w^\tau - \frac{1}{|\mathcal{S}^{\tau}|} \sum_{i\in\mathcal{S}^{\tau}} \big(w^{\tau-1} - \ldots -\frac{1}{|\mathcal{S}^{\tau}|} \sum_{m\in\mathcal{S}^{1}} (w^{0} - \frac{1}{|\mathcal{S}^\tau|} \sum_{l\in\mathcal{S}^{0}} (w^{0} - w_l^{0}) +\\
     &- w^1_m) - ... - w_j^{\tau-1}) -w_i^\tau \big)\Big) 
\end{align}
As in standard SGD, each model implicitly contains information on the previous updates. By unraveling the summation over $\tau$, we get
\begin{align}
    w_{\ours}^{t'+W} &= w^0 - \frac{1}{|\mathcal{S}^0|} \Big(\underbrace{\sum_{i\in\mathcal{S}^0} (w^0 - w_i^0) + ... + \sum_{i\in\mathcal{S}^{t'}} (w^{t'} - w_i^{t'})}_{\textcolor{red}{\tau\leq t'}} + \\
        &+  \underbrace{\underbrace{\frac{W-1}{W}}_{t'+W-(t'+1) = W-1} \sum_{i\in\mathcal{S}^{t'+1}}(w^{t'+1}-w_i^{t'+1}) + ... + \frac{1}{W} \sum_{i\in\mathcal{S}^{t'+W-1}} (w^{t'+W-1} - w_i^{t'+W-1})}_{\textcolor{red}{t'<\tau<t'+W}}\Big)=\\
        &= w^{t'} - \frac{1}{|\mathcal{S}^0|} \Big(\frac{W-1}{W} \sum_{i\in\mathcal{S}^{t'+1}}(w^{t'+1}-w_i^{t'+1}) + ... + \frac{1}{W} \sum_{i\in\mathcal{S}^{t'+W-1}} (w^{t'+W-1} - w_i^{t'+W-1})\Big)=\\
        &= w^{t'} - \frac{1}{|\mathcal{S}^0|} \sum_{\tau=t'+1}^{t'+W-1} \frac{t'+W-\tau}{W} \sum_{i\in\mathcal{S}^\tau} (w^\tau - w_i^\tau).
\end{align}

If we drop the constraint $\frac{N_i}{N} = \frac{1}{|\mathcal{S}^t|}$ and insert the server learning rate $\eta_s$, we can summarize the results as 
\begin{equation}
    w_{\ours}^{t'+W}=w^{t'} - \eta_s \sum_{\tau=t'}^{t'+W-1} \frac{t'+W-\tau}{W} \sum_{i\in\mathcal{S}^\tau} \frac{N_i}{N} (w^\tau - w_i^\tau ),
\end{equation}
obtaining Eq.~\ref{eq:lr_decay}.

\end{document}